# Support Sufficiency as Consequence-Sensitive Compression in Belief Arbitration

Mark Walsh, PharmD
Corresponding email: mark.walsh@rutgers.edu

**Abstract**

When a system commits to a hypothesis, much of the evidential structure behind that commitment is lost to compression. Standard accounts assume that selected content and scalar confidence are sufficient for downstream control. This paper argues that they are not, and that determining what must survive compression is itself a consequence-sensitive problem. We develop a recurrent arbitration architecture in which active constraint fields jointly determine a hypothesis geometry over candidates. Rather than carrying that geometry forward in full, the system compresses it into a support-aware control state whose resolution is regulated by the current consequence geometry, arbitration memory, and resource constraints.

A bounded objective formalizes the resulting tradeoff. Preserving too little support collapses policy-relevant distinctions, producing controllers that select content adequately while misrouting verification, abstention, and recovery. Preserving too much support fragments support-conditioned learning across overly fine contexts, degrading adaptation even as local discrimination improves. These two failure modes, under-resolution and over-partitioning, yield an ordered set of controller-level predictions. A minimal repeated-interaction simulation confirms this ordering. Adaptive controllers that regulate support resolution outperform all fixed-resolution controllers in cumulative utility. Agile adaptive control outperforms sluggish adaptive control. Fixed high-resolution control achieves the best commitment accuracy but still trails both adaptive controllers because resource cost and learning fragmentation offset the gains from richer support retention.

The central conclusion is that support sufficiency should be understood not as a static representational threshold, but as a dynamic compression criterion. Robust arbitration depends on preserving the smallest support structure adequate for policy under the current consequence landscape, and on regulating that structure as conditions change across repeated cycles of inference and action.

## 1. Introduction

Arbitration between competing hypotheses requires compression. A system that acts on what it believes must discard much of the evidential structure that led to that belief. In standard treatments, the products of this compression are a selected content state and, in richer accounts, a scalar confidence estimate. These are often taken to be sufficient for downstream control. The present paper argues that they are not, and that the question of what must survive compression is itself a consequence-sensitive problem.

The insufficiency of content and scalar confidence for robust control has been noted in several adjacent literatures. Metacognitive accounts have shown that confidence can be miscalibrated, dissociated from first-order evidence, or insensitive to the structure of uncertainty (Fleming and Daw 2017; Guo et al. 2017). Decision-theoretic work has emphasized that rational action depends on more than point estimates (Berger 1985). Work on abstention, reject-option classification, and selective verification has shown that withholding commitment can be optimal under conditions that scalar confidence alone may not distinguish (Chow 1970; Pleskac and Busemeyer 2010). A recent proposal developed this concern into an explicit architectural claim, arguing that arbitration requires preserving a structured support code beyond content and confidence. That support code is a compressed summary of why a belief should be trusted, deferred, verified, or revised. That framework, referred to here as Constraint Geometry (CG), introduced a three-regime hierarchy of arbitration and showed that matched content and matched confidence can mask policy-relevant differences in underlying support structure (Walsh 2026b). Our earlier simulation work then showed, in a simpler regime-shift setting, that compact support summaries can improve calibration and alter downstream control even when the content-level decision computation is held fixed (Walsh 2026a).

The present paper takes that claim as a starting point and asks the next question. If support must survive compression, how much of it should survive, at what resolution, and under what conditions? We argue that this is a consequence-sensitive problem. The answer depends not only on the current hypothesis geometry, but also on the local consequence geometry, the cost of preserving finer support, and the system's learned history of how compressed support states have projected into downstream consequences. The local consequence geometry here refers to the structure of available policies and their expected outcomes. On this view, support sufficiency is not a fixed representational threshold. It is a bounded control criterion for recurrent arbitration.

The central claim is that robust arbitration requires regulating support resolution rather than merely preserving more or less support in general. Too little retained support collapses distinctions that matter for policy. Too much retained support can fragment support-conditioned learning and degrade adaptation across time. The problem is therefore not maximal support retention, but consequence-sensitive compression. What is needed is the smallest support structure adequate for control under the current stakes, costs, and dynamics of repeated interaction.

## 1.1 Recurrent architecture and notation

We treat arbitration as a recurrent process inside an evolving agent-environment loop rather than as a one-way progression from representation to action. At each time step, a set of active constraints shapes a geometry over candidate hypotheses. Arbitration does not carry that geometry forward in full. It compresses it into a support-aware control state that is used to select policy within a current consequence geometry. Policy then alters the conditions under which later hypothesis geometry is formed. Support sufficiency is therefore not a property of one-shot representation alone. It is a property of whether compression preserves robust policy-relevant differentiation across repeated cycles of inference and action.

To keep the formalism general, we begin with a time-indexed family of active constraint fields $\Lambda_{i,t}{}_{i=1}^{n_t}$ over hypothesis space. These fields may arise from present input, retained structure from prior experience, or internally generated constraints recruited during ongoing inference. We do not require a fixed inventory of sources. What matters is that, at time $t$, the jointly active fields determine the current geometry of support over candidate hypotheses. We write the resulting hypothesis geometry as $H_t = \mathcal{C}\left(\Lambda_{i,t}{}_{i=1}^{n_t}\right)$, where $\mathcal{C}$ is a generic composition operator. In the Bayesian special case, $H_t$ recovers the posterior geometry induced by prior and likelihood intersection. More generally, $H_t$ denotes the effective geometry of support over hypothesis space, including concentration, multimodality, conflict structure, anisotropy, stability, and other features that may matter for downstream control.

We distinguish this from the current consequence geometry, denoted $\mathcal{Z}_t$. Whereas $H_t$ characterizes how candidate hypotheses are currently supported, $\mathcal{Z}_t$ characterizes the local structure of available policies and their expected consequences. It includes whatever distinctions are relevant to current control, such as asymmetric error costs, verification costs, abstention options, or other features of the present action landscape. Hypothesis geometry and consequence geometry therefore play different functional roles. The former organizes what can be inferred. The latter organizes what is at stake in acting.

Arbitration links these two spaces through a bounded compression. We allow the system to regulate its current support resolution, denoted $\rho_t$, as a function of current hypothesis geometry, arbitration memory, and consequence geometry: $\rho_t = \Gamma(H_t, M_t, \mathcal{Z}_t)$. Here $\Gamma$ is the resolution-control policy. Its role is to determine how finely support must be preserved at the current time in light of what is at stake and what has been learned from prior cycles.

Given the chosen resolution $\rho_t$, the system compresses the current hypothesis geometry into a support code $S_t^{\rho_t} = f_{\rho_t}(H_t)$, and derives a selected content state $X_t = x(H_t)$. The selected content $X_t$ is the committed hypothesis at time $t$. The support code $S_t^{\rho_t}$ preserves a low-dimensional summary of policy-relevant structure in the current hypothesis geometry at the chosen resolution.

Together, the selected content and compressed support code define the current arbitration state $A_t = \left(X_t, S_t^{\rho_t}\right)$. Policy is then selected from the current arbitration state, conditioned by arbitration memory and the current consequence geometry: $\pi_t = \Pi(A_t, M_t; \mathcal{Z}_t)$. We use arbitration memory $M_t$ for the learned state that carries forward consequence-relevant structure across cycles. This extends the earlier audit-trail idea by emphasizing its active role in present policy selection and resolution control rather than retrospective record alone.

After policy is selected and executed, the system encounters an outcome $O_t$, and arbitration memory is updated according to $M_{t+1} = \Phi(M_t, A_t, \pi_t, O_t)$. In a simple case, $\Phi$ may reduce to an outcome-conditioned calibration update. In richer cases, it may update value estimates, trust weights, or support-conditioned routing policies. The common point is that the system learns how compressed arbitration states project into downstream consequences, and this learned relation changes future arbitration.

Putting these pieces together, the recurrent architecture may be written compactly as

$$H_t = \mathcal{C}(\Lambda_{i,t}),$$
$$\rho_t = \Gamma(H_t, M_t, Z_t),$$
$$S_t^{\rho_t} = f_{\rho_t}(H_t), \qquad X_t = x(H_t),$$
$$A_t = (X_t, S_t^{\rho_t})$$
$$\pi_t = \Pi(A_t, M_t; Z_t),$$
$$M_{t+1} = \Phi(M_t, A_t, \pi_t, O_t).$$

This sequence should not be read as a purely feedforward pipeline. It is a single cycle inside a recurrent loop. Current policy helps determine later inferential conditions, and poor compression can therefore degrade not only the present decision but the quality of future inference. Figure 1 summarizes this recurrent consequence-sensitive arbitration architecture.

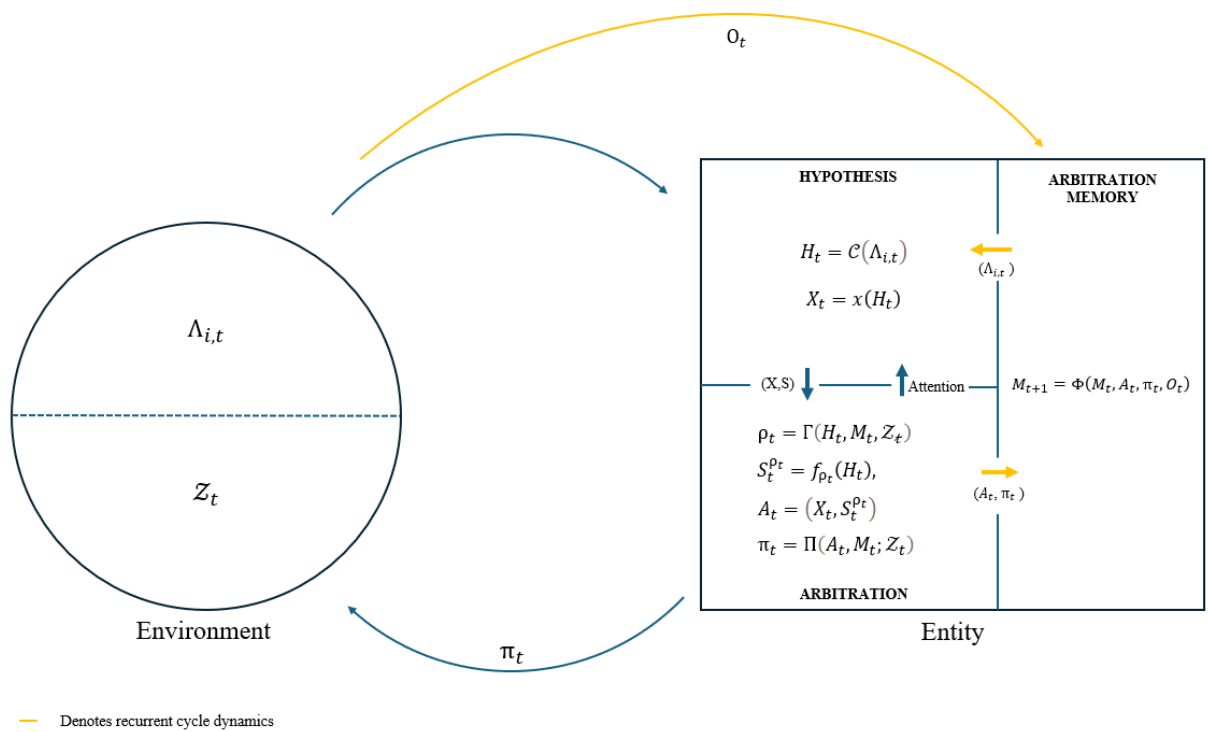

**Figure 1. Recurrent consequence-sensitive arbitration.** Active environmental constraint fields $\Lambda_{i,t}$ shape hypothesis geometry $H_t$, from which selected content $X_t$ and support code $S_t^{\rho_t}$ are derived for arbitration. Here $Z_t$ denotes the idealized current consequence geometry of the environment. Arbitration memory $M_t$ and $Z_t$ jointly regulate support resolution and policy selection. Policy $\pi_t$ acts back on the environment, altering later inferential

conditions. For simplicity, $O_t$ is shown as direct environmental feedback to arbitration memory and denotes the observed post-policy consequence signal used to update $M_t$.

When commitment is deferred rather than immediately executed, we allow arbitration memory to shape further refinement of the current hypothesis geometry: $H_t^{(k+1)} = \Psi\left(H_t^{(k)}, M_t\right)$. Here $k$ indexes within-cycle refinement steps. We include $\Psi$ only to represent the possibility of refinement under deferral.

## 1.2 Single-cycle arbitration and support sufficiency

A single arbitration cycle may be viewed as a cross-section through the recurrent architecture shown in Figure 1. At time $t$, the active constraint fields jointly determine a hypothesis geometry $H_t$. From that geometry, the system derives a selected content state $X_t$ and a compressed support code $S_t^{\rho_t}$. We take as established that matched content and even matched scalar confidence can mask policy-relevant differences in support structure (Walsh 2026b). The issue here is not whether support matters, but what compressed support state is sufficient for control under a given consequence landscape.

The relevant state in the present framework is therefore the arbitration state $A_t = \left(X_t, S_t^{\rho_t}\right)$ which combines selected content with a compressed support code whose current resolution is set by $\rho_t$. The role of $S_t^{\rho_t}$ is not to reproduce the full hypothesis geometry. It is to preserve those features of $H_t$ that matter for downstream control.

This framing overlaps with standard decision-theoretic intuitions, but it addresses a different problem. Standard decision theory typically takes the belief state as given and asks which action maximizes expected utility. Here the central question is upstream of that optimization. We ask what compressed state must survive an arbitration bottleneck for downstream action selection to remain adequate. The target of analysis is therefore not policy selection alone, but the compression map $f_{\rho_t}$ that determines what policy-relevant structure remains available for control.

Let $\pi_t^{\rho_t} = \Pi(A_t, M_t; \mathcal{Z}_t)$ denote the policy chosen from the compressed arbitration state at current support resolution $\rho_t$. We contrast this with an idealized full-geometry policy $\pi_t^* = \Pi^*(H_t, M_t; \mathcal{Z}_t)$, which is the policy that would be chosen if the full current hypothesis geometry were available for downstream control. The purpose of the support code $S_t^{\rho_t}$ is not to reconstruct $H_t$ in full. It is to preserve enough of its structure that the compressed policy $\pi_t^{\rho_t}$ remains appropriately aligned with $\pi_t^*$ for the relevant task and consequence landscape.

This suggests a natural criterion. The current consequence geometry induces an action-relevant partition over hypothesis space. Some distinctions in $H_t$ matter because they place the system on different sides of this partition and therefore lead to different rational policies. Other distinctions do not matter because, despite altering the fine structure of the current hypothesis geometry, they do not change which policy region should be entered. Support sufficiency is the condition that the compression $f_{\rho_t}$ preserves the distinctions that matter and only those distinctions to the extent required for robust control.

In the strongest case, support sufficiency would mean exact preservation of policy at the level of the current cycle. In that idealized form, the requirement would be $\Pi(A_t, M_t; Z_t) = \Pi^*(H_t, M_t; Z_t)$. In practice, exact equality is often stronger than necessary. The more useful criterion is utility-based. Let $U(\pi)$ denote the utility of the policy $\pi$ in the current task context. Then a compressed arbitration state is support-sufficient when the expected utility loss induced by compression remains acceptably small:
$E\big[U(\pi_t^*) - U\big(\pi_t^{\rho_t}\big)\big] \leq \varepsilon$ . The expectation operator averages over the relevant distribution of cases, so this expression measures the expected control loss induced by compression. Compression is sufficient when that loss remains within an acceptable tolerance $\varepsilon$. The criterion is intended as a schema rather than a universally computable rule. Its role is to state what support sufficiency would require in principle, while later sections operationalize that requirement in one limited simulation setting.

In principle, $\varepsilon$ would be set by the steepness of the current consequence geometry and the granularity of the action-relevant partition it induces. Where nearby hypothesis geometries lead to sharply different policy regions, even small compression losses can cross partition boundaries, and $\varepsilon$ must be tight. Where the partition is coarse and the consequence landscape is flat, larger compression losses remain tolerable. The present paper does not derive $\varepsilon$ from first principles, but this dependence on consequence structure is what makes the criterion consequence-sensitive rather than merely approximate.

This criterion also clarifies what the present framework does not claim. The aim is not to maximize retained support. The aim is to preserve policy-relevant differentiation. Support that does not change which policy region should be entered is expendable. Support that does change downstream policy is not. The relevant question is therefore not how much of the current hypothesis geometry can be preserved, but what is the smallest retained structure needed to preserve acceptable policy quality in the current consequence geometry.

### 1.3 Dynamic support resolution and the situational optimum

The cross-space criterion developed above does not imply that support sufficiency is fixed at a single universal resolution. On the contrary, the action-relevant partition induced by consequence geometry can change with stakes, asymmetries, verification value, drift, or other features of the current control problem. The appropriate support resolution is therefore situational. Robust arbitration requires not one globally best compression level, but the ability to preserve more or less support as circumstances change.

We capture this by treating support resolution as a regulated variable, $\rho_t$, rather than as a fixed architectural constant. At time $t$, the system selects a support resolution appropriate to the current hypothesis geometry, current arbitration memory, and current consequence geometry: $\rho_t = \Gamma(H_t, M_t, Z_t)$. The role of $\Gamma$ is to determine how finely policy-relevant structure must be preserved in the present cycle. In low-stakes or routine conditions, coarse compression may be sufficient. In high-stakes, asymmetric-risk, or drift-prone conditions, finer support may be required to preserve distinctions that would otherwise collapse under content and scalar confidence alone.

Support resolution carries both benefits and costs. Increasing $\rho_t$ can improve downstream policy because it preserves distinctions in hypothesis geometry that matter for navigating the current consequence geometry. At the same time, finer support is not free. It incurs resource costs, and it can also over-partition arbitration space in ways that weaken learning. Similar cases that might otherwise contribute to a common learned mapping in arbitration memory may instead be split into increasingly narrow support-conditioned contexts. The result can be slower adaptation, poorer generalization, and greater brittleness under shift.

This objective has a structural parallel to the bounded rationality tradition and formalized more recently as resource-rational analysis, which treats cognitive mechanisms as optimizing expected utility under computational cost constraints (Lieder and Griffiths 2020; Simon 1955). This suggests a bounded objective of the following form:

$$\rho_t^* = \arg\max_{\rho} \; \left[ E\big( U(\pi_t^\rho) \mid Z_t, M_t \big) - \lambda_{\text{res}} \, C_{\text{res}}(\rho) - \lambda_{\text{frag}} \, C_{\text{frag}}(\rho, M_t) \right].$$

Here $E\big( U(\pi_t^\rho) \mid Z_t, M_t \big)$ is the expected utility achieved by policy selected from the compressed arbitration state at resolution $\rho$, $C_{\text{res}}(\rho)$ is the resource cost of sustaining that resolution, and $C_{\text{frag}}(\rho, M_t)$ is the learning cost induced by over-partitioning support-conditioned contexts relative to the current maturity of arbitration memory. The parameters $\lambda_{\text{res}}$ and $\lambda_{\text{frag}}$ weight these costs relative to policy utility.

This objective is schematic rather than universal. Its role is to express the logic of the theory. Increasing support resolution is useful only to the extent that the additional distinctions preserved by $f_\rho$ correspond to genuine differences in the action-relevant partition of consequence space. Once further distinctions no longer alter what should be done, they remain only as cost. Worse, if they fragment support-conditioned learning, they can reduce effective control by weakening the ability of arbitration memory to generalize across cases and adapt under changing conditions.

The same logic can be expressed in loss form. Let $L_{ctrl}(\rho) = E\big[U(\pi_t^*) - U(\pi_t^\rho)\big]$ denote the expected control loss induced by compression at resolution $\rho$. Then support sufficiency is achieved when this loss remains acceptably small while resource and fragmentation costs remain justified. On this view, the goal is not maximal preservation of hypothesis geometry, but minimal acceptable loss of policy differentiation under bounded resources.

This framing makes the two characteristic failure modes transparent. When $\rho_t$ is too low, distinct cases that should lead to different policies collapse into the same arbitration state. The system then under-resolves the current consequence geometry. Verification may be under-allocated, abstention may fail to trigger, and degraded or conflicted support may be treated as ordinary uncertainty. When $\rho_t$ is too high, the opposite problem emerges. The system preserves distinctions that exceed what current control demands or what arbitration memory can learn reliably from available experience. Similar cases are split across too many contexts, slowing adaptation and weakening stable support-conditioned learning.

The theory therefore predicts a situational interior optimum. Optimal arbitration does not preserve the least possible support, nor the most possible support. It preserves the smallest support structure adequate for the current consequence geometry, given present resource constraints and the current state of arbitration memory. Support sufficiency is therefore neither a minimalism thesis nor a maximization thesis. It is a bounded sufficiency thesis. More specifically, arbitration is a consequence-sensitive compression problem in which the amount of support that survives compression is itself a regulated variable.

## 1.4 Arbitration memory and adaptive coupling

The dynamic view of support resolution makes the role of arbitration memory more precise. $M_t$ does not merely record past outcomes. It helps determine what level of compression is currently adequate, how present support should be interpreted, and how effectively the system can maintain robust coupling across repeated cycles of inference and action. In this sense, arbitration memory is not an accessory to the architecture. It is the mechanism by which consequence-sensitive compression becomes adaptive over time.

Its first role is to make support sufficiency history-sensitive. A present support configuration may warrant different decisions depending on what the system has previously learned about similar cases. A coarse support pattern may be sufficient in one regime because prior cycles have established a stable consequence mapping. The same coarse pattern may be inadequate in another regime because drift, spoofing, or altered stakes have made previously negligible distinctions newly important.

Its second role is to preserve robust coupling across cycles. Because arbitration unfolds in a recurrent loop, the effects of current compression do not end with the current policy. Poor compression can lead not only to an inferior present action, but also to a later inferential situation that is harder to interpret or harder to recover from. Conversely, well-regulated compression can improve later inference by routing the system toward verification, abstention, or selective commitment when those are warranted. Arbitration memory is what allows those consequences to accumulate into better future control.

Support sufficiency therefore cannot be evaluated solely within a single cycle. A support code that appears adequate in one isolated decision may prove inadequate across repeated interaction if it systematically discards the distinctions needed to recognize the consequences of earlier compression choices. The relevant question is not only whether compression yields an acceptable policy now, but whether it preserves stable alignment between evolving hypothesis geometry and evolving consequence geometry across cycles in which policy itself helps shape later conditions of inference.

## 2. Consequences of consequence-sensitive compression

The framework developed above implies that arbitration failure cannot be understood solely as inaccurate content selection. Once support resolution is treated as a regulated variable, failure can arise in at least two directions. If support resolution is too low, policy-relevant distinctions collapse into the same arbitration state. If support resolution is too high, support-conditioned learning fragments across overly fine contexts and adaptive control degrades. Because arbitration unfolds inside a recurrent loop, these failures are not confined to the present cycle. They alter later coupling between inference and action. This section derives those consequences and then states the controller-level predictions that follow from them.

### 2.1 Under-resolution as a control failure

The first characteristic failure mode is under-resolution. This occurs when the support code $S_t^{\rho_t}$ is too coarse to preserve distinctions in the current hypothesis geometry that matter for navigating the current consequence geometry. In this regime, compression erases policy-relevant structure. Cases that ought to lead to different policies are mapped into the same arbitration state and are therefore treated as equivalent for control.

Under-resolution need not manifest as poor first-order content selection. Selected content $X_t$ may remain unchanged, and scalar confidence may remain superficially adequate, even while policy becomes inappropriate. A compressed state may still support the right hypothesis in the narrow sense of content choice, yet fail to distinguish whether the system should act immediately, defer, verify, abstain, or route the case through a more conservative policy. The problem is not that the system has forgotten what it believes. It is that it has compressed away why that belief should be treated differently in the present circumstance.

This is the control-level signature of insufficient compression. A coarse arbitration state is inadequate not because it is less faithful in the abstract, but because it collapses distinctions that sit on different sides of the action-relevant partition induced by $Z_t$. Weak but coherent support, sharp inter-source conflict, and degraded support can all converge on the same selected content, yet still warrant different downstream policies. Under-resolution forces those regimes to converge in control when they should diverge.

The observable consequences are therefore distinctive. Verification will be under-allocated relative to actual risk. Abstention or deferment will fail to trigger when it should. Support degradation will be mistaken for ordinary uncertainty. Recovery under shift will be slower because the system lacks the retained support structure needed to recognize that the current case differs from familiar ones in a policy-relevant way. In short, under-resolution produces a controller that can appear competent at content selection while being brittle at arbitration.

The qualitative prediction is straightforward. When stakes rise, when error costs become asymmetric, or when support conditions become unstable, a controller with inadequately low support resolution will misroute control. The wrong problem is treated as the right kind of problem. That is the defining signature of under-resolution.

## 2.2 Over-partitioning as a learning failure

The second characteristic failure mode is the opposite one. Support resolution can also become too fine. In this regime, the system preserves distinctions that exceed what the current consequence geometry demands or what arbitration memory can learn reliably from available experience. The result is over-partitioning.

Over-partitioning differs from under-resolution in an important way. The immediate arbitration state may appear richer and more specific, but that added specificity is not automatically useful. If the extra distinctions do not alter which policy region should be entered, they add no control value. Worse, because arbitration memory is support-conditioned, overly fine support codes can divide experience across too many narrow contexts. Similar cases that ought to reinforce a common mapping in $M_t$ are instead distributed across sparsely populated bins.

This fragmentation weakens learning in three familiar ways. It reduces the amount of experience available for each support-conditioned update. It impairs generalization across nearby cases. It also makes the system more brittle under drift, because small changes in support semantics can scatter experience across new contexts faster than arbitration memory can reorganize around them. The controller is not failing because it retains too little support. It is failing because it cannot use what it retains efficiently enough to maintain stable consequence-sensitive control.

The observable consequences are therefore different from those of under-resolution. The system may show excessive verification, unstable policy thresholds, poor transfer across superficially different but functionally similar cases, and slow recovery after regime change despite preserving large amounts of support detail. More retained support is not automatically better. Retained support must remain learnable, usable, and justified by the current consequence geometry. Once those conditions fail, added detail becomes a liability rather than an advantage.

This yields the complementary qualitative prediction. Controllers operating near the ceiling of support resolution will often appear attractive under acute high-risk conditions, but they will underperform over longer horizons when routine conditions dominate or when adaptation requires generalization across similar cases. Their failure mode is not coarse blindness. It is fragmented learning.

## 2.3 Closed-loop consequences across repeated cycles

The two failure modes above become more consequential once arbitration is viewed as a recurrent process rather than a one-shot decision stage. Because current policy affects later inferential conditions, support-insufficient compression can damage not only present control but future coupling between inference and action.

Under-resolution can do this by routing the system into the wrong region of consequence space and thereby creating later conditions that are more difficult to interpret. A missed verification, premature commitment, or failure to detect degraded support does not end with the immediate mistake. It alters what the system encounters next, including the structure of available evidence and the pattern of later consequences that arbitration memory must learn from. Compression failure thus propagates forward through the loop.

Over-partitioning has a different closed-loop effect. By fragmenting support-conditioned learning, it weakens the system's ability to integrate the consequences of prior action into future control. In recurrent settings, that failure can accumulate. The controller becomes increasingly data-hungry, increasingly slow to adapt, and increasingly prone to local overfitting of support distinctions that do not generalize.

The loop structure therefore sharpens the meaning of support sufficiency. A compression that looks acceptable in a single isolated cycle may still be inadequate if it systematically undermines later inference or later adaptation. Conversely, a compression that appears coarse in the present cycle may still be sufficient if it reliably preserves the distinctions needed for robust coupling over time. What matters is not the richness of one arbitration state viewed in isolation, but whether the retained structure supports good control across repeated interaction.

This distinguishes the present framework from one-shot compression accounts. The question is not merely how much policy quality is lost by compressing a current state. The deeper question is whether compression preserves stable alignment between evolving hypothesis geometry and evolving consequence geometry across cycles in which the system's own policies help shape later conditions of inference.

## 2.4 Fixed and adaptive resolution-control regimes

Treating support resolution as a regulated variable implies that controllers can differ not only in how much support they preserve, but in how well they regulate support resolution as circumstances change. This yields a second major class of predictions at the level of controller type.

A fixed low-resolution controller should perform acceptably in simple or routine conditions, but will fail when consequence geometry sharpens or when support conditions become unstable. Its characteristic weakness will be under-resolution. It will collapse distinct risk regimes into the same arbitration state and therefore misallocate verification, abstention, or caution.

A fixed high-resolution controller should perform better in acute high-risk or highly asymmetric settings, but will pay persistent resource and fragmentation costs in routine conditions. Its characteristic weakness will be over-partitioning. It will preserve more distinctions than current control usually requires, weakening support-conditioned learning and generalization over time.

A fixed midrange controller may outperform either extreme in stable mixed environments, but it remains limited by the fact that the situational optimum is not constant. When stakes rise sharply, it may fail to recruit sufficient support resolution. When the landscape flattens, it may preserve more detail than necessary. Its performance will therefore be acceptable on average while remaining suboptimal in precisely those conditions where resolution control matters most.

Adaptive controllers should differ from fixed controllers in a more interesting way. A sluggish adaptive controller may eventually move toward an appropriate support resolution, but with a lag that makes it vulnerable to abrupt changes in stakes, drift, or support degradation. By contrast, an agile adaptive controller should increase or decrease support resolution as the consequence geometry changes, preserving finer distinctions when the current control problem becomes steep and relaxing them when coarse compression is again sufficient.

The comparative prediction is therefore ordered rather than binary. The most effective controller is not the one that preserves the most support, nor the one that preserves the least, but the one that best regulates support resolution in a consequence-sensitive manner. In the simulation developed below, that prediction becomes concrete. Adaptive controllers should outperform fixed controllers in cumulative utility, and agile adaptive control should outperform sluggish adaptive control when changing conditions reward fast reallocation of support resolution. High fixed resolution may still improve local commitment accuracy, but need not yield the best overall arbitration once resource cost and learning fragmentation are included.

## 3. Minimal simulation of consequence-sensitive arbitration

To test the framework in a controlled setting, we implemented a minimal repeated-interaction simulation in which arbitration compresses current hypothesis geometry into a support-aware control state under changing consequence geometry. The goal of the simulation was not to model any one biological or engineered system in detail. It was to test the central comparative claim developed in Section 2. In particular, we asked whether controllers that regulate support resolution in a consequence-sensitive manner outperform controllers with poorly matched fixed resolutions, and whether preserving more support always improves arbitration once resource costs and support-conditioned learning are taken into account.

The simulation was intentionally modest. It used a binary latent state, a small set of evidence channels, a compact support vocabulary, and a small action set. This kept the setting simple enough to interpret while still allowing the dissociations required by the theory to appear. The basic design was continuous with the Gaussian observation-style constructions used in our earlier calibration work, but extended them to include dynamic support compression, variable consequence geometry, and explicit controller comparisons (Walsh 2026a).

## 3.1 Task structure

At each time step $t$, the environment instantiated a latent binary state $X_t \in 0,1$. The agent did not observe $X_t$ directly. Instead, it received two initial evidence channels, $A$ and $B$, and could optionally recruit a third verification channel, $C$, if policy selected verification. Each channel was generated from a Gaussian observation model of the form $y_{i,t} \mid X_t = x \sim \mathcal{N}\left(x, \sigma_{i,t}^2\right)$. This observation structure follows the standard signal detection framework (Green and Swets 1966). Channel $A$ was stable and moderately reliable. Channel $B$ was variable, alternating between good and degraded conditions. Channel $C$ was available only under verification and was set to be highly reliable.

The environment also varied along two independent regime dimensions. The first was a consequence regime, $Z_t \in \{routine, threat\}$, which determined the current stakes. The second was a support regime, $R_t \in \{stable, shifted\}$, which determined how informative a coarse quality cue about channel $B$ was. This separation was important because it let us distinguish changes in what was at stake from changes in how support should be interpreted. The current paper predicts that both matter, but in different ways.

From the initial evidence channels, the agent formed a provisional hypothesis state and a compact support vector $g_t = (m_t, c_t, b_t)$, where $m_t$ marked low-margin cases, $c_t$ marked conflict between channels, and $b_t$ marked a bad-quality cue. These three descriptors were chosen because they were enough to instantiate the key policy-relevant distinctions developed in the theory which were coherent but weak support, inter-channel conflict, and degraded support conditions. In this sense, the simulation does not attempt to preserve the full upstream support geometry. It preserves only a small vocabulary of distinctions that can change what the rational controller should do.

The geometric vocabulary of the theory is therefore instantiated here in a deliberately minimal discrete form. The three-bit support vector does not exhibit the continuous, high-dimensional structure that 'hypothesis geometry' invokes in the general framework. That gap is intentional. The simulation tests whether the consequence-sensitive compression logic holds even in a setting where geometry reduces to a small combinatorial structure. Richer continuous instantiations, in which support resolution regulates genuine geometric features such as curvature, anisotropy, or manifold structure in hypothesis space, are a natural target for future work.

## 3.2 Support compression and controller classes

Support resolution determined how much of the raw support vector remained available to arbitration. We used three discrete levels. At low resolution, no explicit support structure was preserved. At mid resolution, the three support descriptors were collapsed into a single suspiciousness bit. At high resolution, the full three-bit vector was retained. This gave the simulation a direct operationalization of the paper's core compression claim. Controllers differed not in access to different worlds, but in how much of the same raw support structure survived the arbitration bottleneck.

The agent's action set was {act, verify, abstain}. Acting committed to the current provisional decision. Verification incurred an immediate cost but revealed an additional reliable evidence channel before commitment. Abstention withheld commitment at a regime-dependent penalty. These options were chosen because they correspond directly to the policy distinctions the theory claims support compression must preserve. A support code is useful here only if it helps route cases among these control options more appropriately under changing stakes.

We compared five controller classes. The first three were fixed-resolution controllers that always operated at low, mid, or high support resolution. The fourth was a sluggish adaptive controller. The fifth was an agile adaptive controller. In the present simulation, the adaptive controllers were idealized benchmarks in the sense that they were informed about the current consequence regime $Z_t$, but differed in how quickly they adjusted support resolution. The agile controller updated immediately to the desired resolution for the current case. The sluggish controller updated only gradually through an inertia rule. This distinction let us isolate the value of consequence-sensitive regulation itself before taking on the harder problem of learning the resolution-control policy $\Gamma$ from experience.

### 3.3 Learning and evaluation

Arbitration memory $M_t$ was implemented as support-conditioned action values over the current compressed policy state. Action selection was $\varepsilon$-greedy, and value updates used an exponential moving average (Watkins and Dayan 1992). This simple Q-style implementation was sufficient for the present purpose because it naturally generated the fragmentation effect described in Section 2. As support-conditioned state spaces became finer, experience was spread across more contexts and learning slowed. That made the cost of over-partitioning operational rather than merely conceptual.

Per-trial utility combined task payoff and support-resolution cost. Correct commitment yielded positive utility. Incorrect commitment incurred penalties that depended on the consequence regime, with misses in the threat regime penalized more heavily than false positives. Verification incurred a fixed cost. Abstention incurred a smaller but still nonzero penalty. Higher support resolution also carried an explicit resource cost. Before running the full experiment, we performed a small sanity pass on these parameter choices to ensure that act, verify, and abstain were all genuine tradeoffs rather than dominated options. In the final configuration, clean cases generally favored direct action, moderately suspicious cases often favored verification, and the most extreme suspicious cases could justify abstention under threat. This mattered because the simulation needed a live policy space in order to test the paper's claims.

We evaluated each controller over 50,000 trials and averaged results across 10 random seeds. The primary outcome was cumulative utility. Secondary metrics included verification allocation, abstention allocation, commitment accuracy, calibration error, support efficiency, and adaptation around regime shifts. This evaluation strategy reflects the logic of the framework. The primary target is not maximal local discrimination in isolation. It is the quality of consequence-sensitive arbitration over repeated interaction.

### 3.4 Results

The results supported the central comparative claim of the paper. Table 1 summarizes the main performance metrics across controllers. The most important comparative result is shown in Figure 2, which plots cumulative utility across seeds. The best-performing controller was the agile adaptive controller, with mean cumulative utility of approximately 17,468 across seeds. The sluggish adaptive controller performed next, at approximately 17,017. Among the fixed controllers, the high-resolution controller outperformed the mid- and low-resolution controllers on cumulative utility, but still trailed both adaptive controllers, at approximately 16,177. The mid-resolution controller followed closely at approximately 16,016, and the low-resolution controller performed worst overall at approximately 14,858. This ordering matters because it matches the prediction of Section 2.4. The best controller was not the one that preserved the most support or the least. It was the one that best regulated support resolution as consequence geometry changed.

| Controller | Cumulative Utility (mean +/- SD) | Commitment Accuracy | ECE | Verification Rate Routine | Verification Rate Threat | Mean ρ |
|---|---|---|---|---|---|---|
| Adaptive Agile | 17,467 ± 1,231 | 0.8602 | 0.057 | 0.2394 | 0.3057 | 1.08 |
| Adaptive Slow | 17,017 ± 1,420 | 0.8571 | 0.058 | 0.2426 | 0.3062 | 1.19 |
| Fixed High | 16,176 ± 969 | 0.8725 | 0.055 | 0.2697 | 0.3148 | 2.00 |
| Fixed Mid | 16,016 ± 1,262 | 0.8521 | 0.059 | 0.2596 | 0.2557 | 1.00 |
| Fixed Low | 14,858 ± 1,720 | 0.8394 | 0.065 | 0.2093 | 0.2913 | 0.00 |

**Table 1. Controller-level performance summary across 10 simulation seeds.** Values report cumulative utility, commitment accuracy, calibration error, regime-specific verification rates, and mean support resolution. Adaptive controllers achieve the highest cumulative utility despite not preserving the highest average support resolution, consistent with consequence-sensitive regulation rather than maximal retention of support.

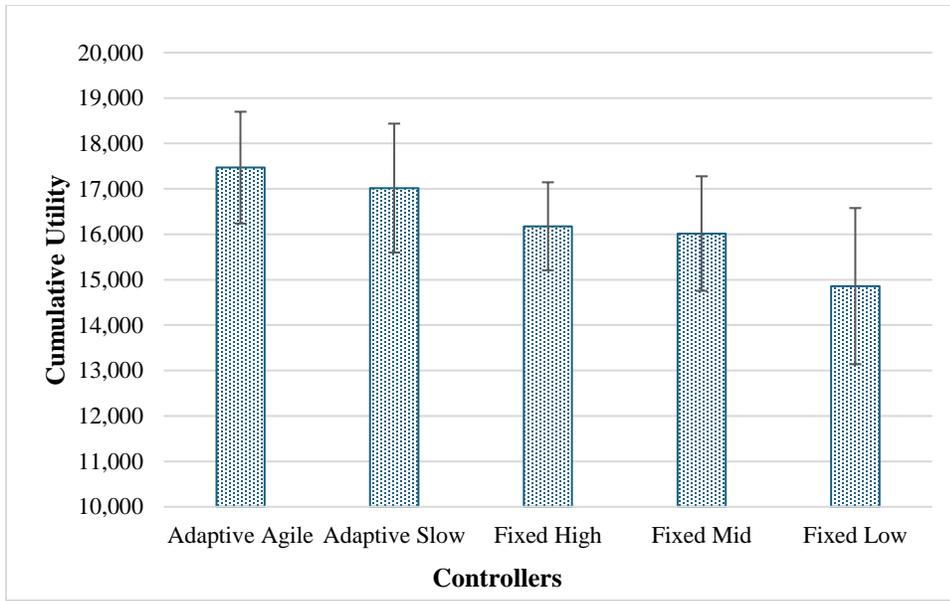

**Figure 2. Cumulative utility by controller.** Bars show mean cumulative utility across 10 seeds; error bars denote standard deviation. Adaptive controllers outperform fixed-resolution controllers, and the agile adaptive controller performs best overall.

The gap between the agile and sluggish adaptive controllers is also informative. Both had access to the same idealized information about current stakes, yet the slower controller still underperformed because delayed support-resolution adjustment imposed real control costs. This is exactly the dissociation the dynamic-resolution framework predicts. The relevant competence is not merely access to support, but the ability to regulate support resolution at the right time.

The fixed controllers clarify the two failure modes developed in Section 2. The high-resolution controller achieved the best commitment accuracy overall, at approximately 0.872, compared with 0.852 for the fixed mid controller and 0.839 for the fixed low controller. Yet this gain in accuracy did not translate into the highest cumulative utility. The high-resolution controller paid for its richer support retention through persistent resource cost and more fragmented support-conditioned learning. This is a direct instantiation of the over-partitioning argument. More retained support improved local discrimination, but did not yield the best overall arbitration once resource and learning costs were included.

The low-resolution controller showed the complementary failure mode. Its low average support resolution reduced resource cost, but it systematically under-resolved policy-relevant distinctions. This appeared in lower cumulative utility, lower commitment accuracy, and the worst calibration among the controllers, with expected calibration error of about 0.065 versus roughly 0.055 to 0.059 for the others. In other words, the low-resolution controller did not fail simply because it retained less information. It failed because it collapsed cases that should have been routed differently for control.

The regime-specific policy allocations were also consistent with the theory. The adaptive controllers verified more often under threat than under routine conditions, whereas the fixed controllers were less well matched to the current consequence regime. This supports the claim that consequence-sensitive arbitration is not simply a matter of richer representation, but of preserving and using the right distinctions when the stakes demand them. The adaptive controllers did not win by verifying indiscriminately. They won by reallocating support resolution and verification more appropriately as conditions changed. Figure 3 shows that the adaptive controllers reallocated verification more appropriately between routine and threat conditions.

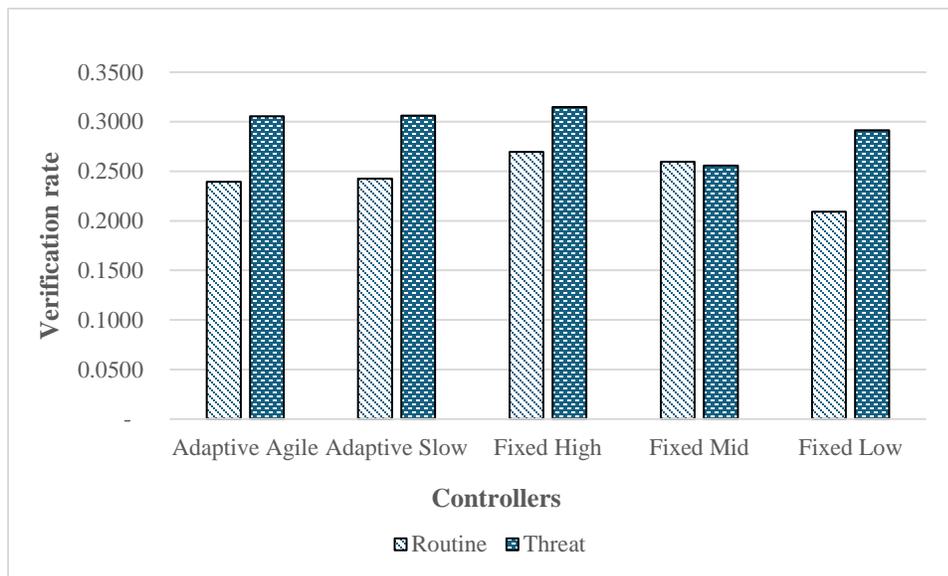

**Figure 3. Verification allocation in routine and threat regimes.** Grouped bars show mean verification rates by controller across routine and threat conditions. Adaptive controllers verify more in threat than in routine conditions, consistent with consequence-sensitive reallocation of arbitration under changing stakes.

Calibration results strengthened this interpretation. The fixed low-resolution controller had the worst expected calibration error, while the higher-resolution and adaptive controllers showed better alignment between commitment and success. This provides useful continuity with the broader research program. The present simulation was not built primarily as a calibration study, but it nevertheless showed that support-aware arbitration can improve not only utility but also calibration under changing support conditions.

Taken together, these results support three conclusions. First, under-resolution is a real and costly failure mode. Second, preserving more support is not itself sufficient for robust arbitration. Third, dynamic regulation of support resolution provides a genuine advantage over poorly matched fixed-resolution control.

### 3.5 Interpretation and limits

These results are meaningful, but the level of claim should remain modest. The simulation demonstrates the value of consequence-sensitive regulation of support resolution in a minimal setting. It does not yet show that agents can learn the full resolution-control policy Γ from experience, since the adaptive controllers were idealized and oracle-informed with respect to current stakes. Nor does it exhaust the richer possibilities suggested by the broader framework, such as dimension-specific support resolution or more elaborate within-cycle refinement. The present role of the simulation is narrower. It provides a proof of concept that the theoretical distinction drawn in Section 2.4 is operational and consequential.

That proof of concept is already substantial. The simulation shows that the central contrast of the paper is not merely between low and high information retention, but between poorly matched fixed compression and consequence-sensitive control of compression. It also shows that the costs of over-retaining support are not merely conceptual. They can appear concretely as fragmented learning, weaker support efficiency, and lower overall utility even when immediate accuracy improves.

For present purposes, that is enough. The simulation does not settle the broader theory in full richness, but it does make the comparative claim operational. In a minimal setting, consequence-sensitive regulation of support resolution outperformed poorly matched fixed-resolution control, and did so in exactly the pattern the framework predicts.

## 4. Relation to adjacent frameworks

The distinctive claim of this paper is that arbitration must be understood as a consequence-sensitive compression problem. The system must preserve enough structure from current hypothesis geometry to sustain adequate policy under current consequence geometry, while avoiding unnecessary resource cost and fragmentation of support-conditioned learning. The simulation sharpens this claim by showing that adaptive regulation of support resolution can outperform both low-resolution and high-resolution fixed control, and can do so even when higher fixed resolution improves local commitment accuracy. Section 4 locates that claim relative to adjacent frameworks by asking a more specific question in each case. Not whether uncertainty matters, not whether action should be consequence-sensitive, and not whether compression under constraints is important. Rather, what structure must survive the arbitration bottleneck, and at what resolution, for policy to remain adequate across repeated cycles of inference and action.

### 4.1 Confidence and metacognitive calibration

The most immediate point of contact is with work on confidence and metacognitive calibration. Those traditions ask how systems estimate uncertainty, how confidence relates to first-order evidence, and how calibration can be improved when confidence is misaligned with accuracy (Fleming and Daw 2017; Guo et al. 2017; Maniscalco and Lau 2012). We share those concerns, but the present framework departs from a scalar view of uncertainty. Its starting point is that content and scalar confidence can be jointly insufficient for control because distinct support geometries may warrant different policies even when selected content and apparent confidence are similar (Walsh 2026b). A scalar confidence signal may indicate that a case is uncertain, but it need not distinguish whether that uncertainty arises from weak but coherent support, sharp conflict among constraints, degraded support pathways, or changed support semantics under drift. Computational accounts of confidence have converged on a similar concern, showing that the neural representations underlying confidence carry structure beyond a single scalar estimate (Kepecs and Mainen 2012; Pouget et al. 2016).

That difference matters because calibration alone does not specify what support structure must survive compression for downstream control. Confidence can remain one useful coarse summary of current hypothesis geometry, and improved calibration can be one consequence of better arbitration. But the target here is broader. We ask what additional support structure must be preserved if action, verification, abstention, and recovery under changing conditions are to remain adequate. In that sense, the framework extends metacognitive and calibration-based approaches by shifting the question from uncertainty estimation alone to support-aware control under bottlenecks.

## 4.2 Decision theory, information bottleneck, rate-distortion, and active inference

The framework also overlaps with standard decision theory, but addresses a different stage of the problem. Decision theory typically takes a belief state as given and asks which action maximizes expected utility (Berger 1985). The present paper asks an upstream question. What compressed state must survive an arbitration bottleneck for downstream action optimization to remain adequate at all. The relevant object of analysis is therefore not only the policy map $\Pi$, but also the compression map $f_{\rho_t}$ and the resolution-control policy $\Gamma$ that determine how much support structure remains available for control.

This places the framework near information bottleneck and rate-distortion approaches, but with a different distortion criterion (Tishby et al. 2000; Shannon 1959; Sims 2003). In standard compression settings, distortion is usually defined in terms of reconstruction fidelity or retained mutual information. Here the relevant distortion is consequence-sensitive. What matters is not faithful reconstruction of the full hypothesis geometry as such, but preservation of the distinctions that keep the compressed policy acceptably aligned with the idealized full-geometry policy. The utility-based sufficiency criterion therefore functions as a policy-sensitive distortion bound rather than a generic fidelity constraint. Related work on bounded rational decision-making has shown that information-processing constraints can give rise to hierarchical abstraction and selective compression in multi-step systems, but with distortion defined in terms of KL-divergence from a prior policy rather than consequence-sensitive policy loss (Genewein et al. 2015). Put differently, the relevant loss is policy loss from compression, not representational loss in the abstract. That is not a rhetorical difference. It is the theoretical center of the paper.

The relation to active inference should also be stated in these terms. Active inference explains policy selection under a generative model in terms of expected free energy and precision-weighted inference (Friston et al. 2017; Da Costa et al. 2020). The present paper does not compete with that machinery at the level of global objective or inference-action coupling. It asks a different question that active inference does not itself answer. Given a current hypothesis geometry and a current consequence landscape, what support structure must survive compression into an arbitration state if policy is to remain adequate under access limits, resource limits, and repeated interaction. The present contribution is therefore not a narrower local variant of active inference. It is a bottleneck theory about what must remain available between hypothesis space and policy space. This includes conflict structure, degradation markers, provenance sensitivity, and drift in the meaning of support cues, together with the claim that retaining too much support can also impair control by fragmenting support-conditioned learning.

Resolution control should also be distinguished from precision-weighting in predictive coding. Precision-weighting modulates the influence of prediction errors during inference within a generative hierarchy (Feldman and Friston 2010). By contrast, the regulated resolution variable $\rho_t$ determines what structure survives compression once current hypothesis geometry is projected into an arbitration state for policy selection. The present concern is therefore not gain control within inference, but control over the bottleneck between inference and action.

### 4.3 Abstention, verification, and control under uncertainty

A further point of contact lies in work on abstention, reject-option classification, active information gathering, and related forms of control under uncertainty (Chow 1970; Geifman and El-Yaniv 2017; Pleskac and Busemeyer 2010). These traditions ask when systems should act, defer, verify, or withhold commitment. The present framework aligns naturally with that problem space, but adds a prior architectural question. What support distinctions must survive compression for those control options to be allocated appropriately in the first place.

That question matters because a system can fail at abstention or verification even when its content estimates remain broadly adequate. If the compressed arbitration state does not preserve the structure that distinguishes ordinary uncertainty from degraded support, conflict, or shift, then the controller may systematically misallocate verification or fail to withhold commitment when it should. From this perspective, verification and abstention are not optional add-ons to belief selection. They are part of what defines support sufficiency. A support code is sufficient only if it preserves the distinctions needed to route the current case into the appropriate control region. The simulation makes that point concrete. The adaptive controllers did not win by verifying indiscriminately. They won by reallocating verification more appropriately across consequence regimes.

### 4.4 Drift, support semantics, and adaptive control

The framework also connects closely to work on distribution shift, concept drift, support degradation, provenance, and adaptive control (Gama et al. 2014; Ovadia et al. 2019). These literatures emphasize that conditions under which a system learned may not remain stable, and that robust performance often depends on detecting and adapting to changed regimes. We agree, but the present framework adds a compression-theoretic constraint. Adaptation under drift depends not only on whether the system updates, but on whether compression has preserved the distinctions needed to recognize that a new case falls into a different control regime.

This is where arbitration memory becomes central. Because $M_t$ stores learned consequence mappings across repeated cycles, the adequacy of compression is history-sensitive. A coarse support distinction that was once sufficient may become inadequate after shift. Conversely, a distinction that initially required finer preservation may later become compressible once arbitration memory has learned a stable mapping from coarse support patterns to consequence. The resulting picture is neither static compression nor unconstrained adaptation. It is adaptive consequence-sensitive compression under changing support semantics.

This point also helps explain why the paper places equal emphasis on under-resolution and over-partitioning. Under drift, too little retained support impairs recognition of changed conditions. Too much retained support can fragment learning and slow recovery. Robust arbitration therefore depends not simply on retaining more support, but on retaining support at the right resolution for present control and present learnability. The simulation supports exactly this interpretation. The agile adaptive controller outperformed the sluggish adaptive controller, and both outperformed the fixed controllers, while the fixed high-resolution controller still trailed the adaptive controllers despite preserving the greatest average support and achieving the best local commitment accuracy. The empirical moral matches the conceptual one. Support preservation is not enough on its own. Instead, what matters is consequence-sensitive regulation of support resolution across changing conditions.

Taken together, these comparisons clarify the paper's intended contribution. The framework agrees with metacognitive work that uncertainty matters, with decision theory that action should be consequence-sensitive, with information-theoretic work that compression under constraints is central, and with adaptive-control work that drift makes static solutions brittle. Its distinctive contribution is to join these insights at the arbitration bottleneck and to treat support preservation itself as a regulated variable. The central question is not only what to believe, how confident to be, or which action to choose from a full state. It is what structure must survive compression, and at what resolution, for consequence-sensitive control to remain adequate across repeated cycles of inference and action.

## 5. Discussion

The present paper argued that arbitration should be understood as a consequence-sensitive compression problem. The core claim is not simply that support matters beyond content and scalar confidence. It is that support sufficiency is dynamic. Whether a compressed arbitration state is sufficient depends on whether it preserves the distinctions needed for adequate policy under the current consequence geometry, without incurring unnecessary resource cost or fragmenting support-conditioned learning. Robust arbitration is therefore a problem of regulating support resolution across repeated cycles of inference and action.

The simulation made that comparative claim operational. Controllers with fixed support resolution exhibited the two failure modes predicted by the theory. Low-resolution control under-resolved policy-relevant distinctions and performed worst overall. High-resolution control improved local commitment accuracy, but still underperformed adaptive controllers in cumulative utility because it paid persistent resolution cost and fragmented support-conditioned learning. The adaptive controllers performed best, and the agile adaptive controller outperformed the sluggish one. The most effective controller was therefore not the one that preserved the most support, nor the one that preserved the least. It was the one that best regulated support resolution in a consequence-sensitive manner.

This result sharpens what the present paper adds beyond earlier support-aware formulations. CG argued that support-aware arbitration can preserve policy-relevant structure beyond content and scalar confidence, and the audited calibration model showed that compact support summaries can alter downstream control under regime shift. The present paper extends that line of work by asking a different question. If support must survive compression, how much of it should survive, at what resolution, and when should that resolution change. The answer offered here is bounded and situational. Compression is good enough not when it maximizes fidelity to the full hypothesis geometry, but when it preserves the distinctions needed to enter the right policy region without incurring unnecessary resource cost or fragmentation of learning.

The recurrent structure of the framework is part of why this matters. Support sufficiency should not be judged solely within a single arbitration cycle because current policy affects later inferential conditions. Poor compression can therefore damage not only present control but also later coupling between inference and action. This is one reason the simulation compared controllers over repeated interaction rather than single-shot decisions. The theory is ultimately about maintaining adequate alignment between evolving hypothesis geometry and evolving consequence geometry across time.

A useful way to think about competence in this framework is not as maximal vigilance and not as maximal automation, but as selective control over support resolution. A novice controller often preserves too much support too inefficiently, or too little support too indiscriminately. An expert controller preserves less support in routine conditions, but is better at selectively increasing support resolution when consequence geometry sharpens. The difference is not simply more or less information. It is better regulation of how much information must remain explicit for policy to stay adequate. This helps explain why routine efficiency can coexist with rapid upshifting under changed conditions, and why brittle failure can occur when an otherwise efficient controller does not increase resolution quickly enough under shift.

The same framework may also offer a useful descriptive vocabulary for pathological arbitration, provided the level of claim remains modest. Complex clinical syndromes should not be reduced to single formal failure modes. Even so, the present framework suggests that control can fail in more than one direction. Some systems may fail because they collapse distinctions that should be preserved. Others may fail because they preserve distinctions that no longer pay for themselves in control. Still others may fail because support resolution is regulated well on average but too slowly under abrupt change. Anxious hypervigilance is one everyday illustration. In such cases, threat-relevant support may remain live or be weighted too strongly, so that control becomes inefficient even when the system appears highly vigilant. In the present terms, this looks like dysregulated resolution control in which support is preserved or sampled at a level poorly matched to the current consequence geometry.

The paper remains limited in several important ways. The simulation is intentionally minimal and uses oracle-informed adaptive controllers as idealized benchmarks for consequence-sensitive regulation of support resolution. It therefore does not yet show how agents learn the resolution-control policy $\Gamma$ from experience. Nor does it address richer forms of support resolution, such as dimension-specific sharpening or suppression. The simulation also retains only a modest within-cycle refinement structure.

A further limitation deserves separate attention. The present formalism treats consequence geometry $Z_t$ as available to the resolution-control policy $\Gamma$. In practice, consequence geometry must itself be inferred from available constraint fields and prior outcome history. This introduces a further source of distortion. A system that regulates support resolution against an inaccurate model of current stakes may compress appropriately relative to that model while remaining poorly matched to the actual control problem. The simulation isolates the value of consequence-sensitive regulation under idealized access to current stakes. The further question of how consequence geometry is itself modeled, and how distortion in that model propagates into resolution control, is not addressed here. It connects to a broader problem in which hypothesis geometry, consequence geometry, and outcome signals are all parsed from a common world model rather than arriving as independent inputs. That problem belongs to a later stage of the program.

These are real limitations, but they do not undermine the present result. The point of the simulation was narrower. It shows that the comparative claim of the theory is operational, and that consequence-sensitive control of support resolution can outperform poorly matched fixed-resolution strategies in a minimal setting.

A final architectural point should also be kept in view. The formalism retained the possibility that arbitration memory can participate not only in policy selection across cycles, but also in within-cycle refinement when commitment is deferred. In the present paper, that operator was included only to preserve the architecture of deferred refinement. Its broader implications were not developed here. Even so, its presence matters. It marks the possibility that arbitration may shape not only how a fixed hypothesis state is used, but also how the hypothesis state itself is refined when commitment is withheld. That extension belongs to a broader line of inquiry than the present one, but it is architecturally continuous with the consequence-sensitive arbitration framework developed here.

Taken together, the paper supports a bounded sufficiency view of arbitration. Robust control requires preserving the distinctions that matter now, together with adaptive memory for which distinctions continue to matter across time. On this view, the central problem of arbitration is not merely what to believe or which action to choose, but how much of current hypothesis geometry must remain available for policy in an evolving world.

## 6. Conclusion

We have argued that arbitration should be understood as a consequence-sensitive compression problem. The central issue is not only which hypothesis is selected, but what structure from current hypothesis geometry must survive compression for downstream policy to remain adequate under the current consequence geometry. This reframes support sufficiency as a bounded, dynamic, and consequence-sensitive principle rather than a demand for maximal retention of support.

The paper's comparative claim was made operational in a minimal repeated-interaction simulation. Fixed low-resolution control under-resolved policy-relevant distinctions. Fixed high-resolution control improved local commitment accuracy but still underperformed adaptive controllers in cumulative utility because it paid persistent resolution cost and fragmented support-conditioned learning. The best performance came from consequence-sensitive regulation of support resolution.

The broader implication is that the problem of arbitration is not exhausted by belief, confidence, or action considered separately. It is a problem of what structure must remain available for policy in an evolving world, and how that structure should be regulated across repeated cycles of inference and action. In that sense, the paper's main contribution is not a new uncertainty taxonomy or a new decision rule. It is a bottleneck-level claim about the conditions under which compressed support remains sufficient for robust control.

This also clarifies the broader trajectory of the support-sufficiency program. CG established that support-aware arbitration can preserve policy-relevant structure beyond content and scalar confidence. The audited calibration model showed that compact support summaries can alter calibration and information-seeking under shift. The present paper adds that support preservation is itself a regulated variable. Robust arbitration depends not only on retaining support, but on retaining it at the right resolution for present stakes, present learnability, and future adaptation.

Future work can now take this framework in several directions. One is to learn the resolution-control policy $\Gamma$ rather than idealize it. Another is to move from scalar support resolution to dimension-specific regulation. A third is to test whether richer support codes can be learned under partial observability without fragmenting support-conditioned learning beyond usefulness. These next steps matter, but they do not change the present conclusion. Support sufficiency is best understood not as maximal information retention, but as consequence-sensitive control over what structure remains available for policy.

**References:**

Berger, J. O. 1985. *Statistical Decision Theory and Bayesian Analysis*. 2nd ed. Springer. http://dx.doi.org/10.1007/978-1-4757-4286-2.

Chow, C. 1970. "On Optimum Recognition Error and Reject Tradeoff." *IEEE Transactions on Information Theory* 16 (1): 41–46. https://doi.org/10.1109/TIT.1970.1054406.

Da Costa, Lancelot, Thomas Parr, Noor Sajid, Sebastijan Veselic, Victorita Neacsu, and Karl Friston. 2020. "Active Inference on Discrete State-Spaces: A Synthesis." *Journal of Mathematical Psychology* 99 (December): 102447. https://doi.org/10.1016/j.jmp.2020.102447.

Feldman, Harriet, and Karl J. Friston. 2010. "Attention, Uncertainty, and Free-Energy." *Frontiers in Human Neuroscience* (Switzerland) 4: 215. https://doi.org/10.3389/fnhum.2010.00215.

Fleming, Stephen M., and Nathaniel D. Daw. 2017. "Self-Evaluation of Decision-Making: A General Bayesian Framework for Metacognitive Computation." *Psychological Review* (United States) 124 (1): 91–114. https://doi.org/10.1037/rev0000045.

Friston, Karl, Thomas FitzGerald, Francesco Rigoli, Philipp Schwartenbeck, and Giovanni Pezzulo. 2017. "Active Inference: A Process Theory." *Neural Computation* 29 (1): 1–49. https://doi.org/10.1162/NECO_a_00912.

Gama, João, Indrė Žliobaitė, Albert Bifet, Mykola Pechenizkiy, and Hamid Bouchachia. 2014. "A Survey on Concept Drift Adaptation." *ACM Computing Surveys (CSUR)* 46 (April). https://doi.org/10.1145/2523813.

Geifman, Yonatan, and Ran El-Yaniv. 2017. "Selective Classification for Deep Neural Networks." https://arxiv.org/abs/1705.08500.

Genewein, Tim, Felix Leibfried, Jordi Grau-Moya, and Daniel Alexander Braun. 2015. "Bounded Rationality, Abstraction, and Hierarchical Decision-Making: An Information-Theoretic Optimality Principle." *Frontiers in Robotics and AI* Volume 2-2015. https://doi.org/10.3389/frobt.2015.00027.

Green, D. M., and J. A. Swets. 1966. *Signal Detection Theory and Psychophysics*. Wiley.

Guo, Chuan, Geoff Pleiss, Yu Sun, and Kilian Q. Weinberger. 2017. "On Calibration of Modern Neural Networks." https://arxiv.org/abs/1706.04599.

Kepecs, Adam, and Zachary F. Mainen. 2012. "A Computational Framework for the Study of Confidence in Humans and Animals." *Philosophical Transactions of the Royal Society B: Biological Sciences* 367 (1594): 1322–37. https://doi.org/10.1098/rstb.2012.0037.

Lieder, Falk, and Thomas L. Griffiths. 2020. "Resource-Rational Analysis: Understanding Human Cognition as the Optimal Use of Limited Computational Resources." *Behavioral and Brain Sciences* 43: e1. Cambridge Core. https://doi.org/10.1017/S0140525X1900061X.

Maniscalco, Brian, and Hakwan Lau. 2012. "A Signal Detection Theoretic Approach for Estimating Metacognitive Sensitivity from Confidence Ratings." *Consciousness and Cognition* (United States) 21 (1): 422–30. https://doi.org/10.1016/j.concog.2011.09.021.

Ovadia, Yaniv, Emily Fertig, Jie Ren, et al. 2019. "Can You Trust Your Model's Uncertainty? Evaluating Predictive Uncertainty Under Dataset Shift." https://arxiv.org/abs/1906.02530.

Pleskac, Timothy J., and Jerome R. Busemeyer. 2010. "Two-Stage Dynamic Signal Detection: A Theory of Choice, Decision Time, and Confidence." *Psychological Review* (United States) 117 (3): 864–901. https://doi.org/10.1037/a0019737.

Pouget, Alexandre, Jan Drugowitsch, and Adam Kepecs. 2016. "Confidence and Certainty: Distinct Probabilistic Quantities for Different Goals." *Nature Neuroscience* 19 (3): 366–74. https://doi.org/10.1038/nn.4240.

Shannon, Claude E. 1959. "Coding Theorems for a Discrete Source With a Fidelity Criterion." *International Convention Record*, 142–63.

Simon, Herbert A. 1955. "A Behavioral Model of Rational Choice." *The Quarterly Journal of Economics* 69 (1): 99–118. https://doi.org/10.2307/1884852.

Sims, Christopher A. 2003. "Implications of Rational Inattention." *Journal of Monetary Economics* 50 (3): 665–90. https://doi.org/10.1016/S0304-3932(03)00029-1.

Tishby, Naftali, Fernando C. Pereira, and William Bialek. 2000. "The Information Bottleneck Method." https://arxiv.org/abs/physics/0004057.

Walsh, Mark. 2026a. "Audited Calibration under Regime Shift as a Computational Test of Support-Structured Broadcast." https://arxiv.org/abs/2602.23382.

Walsh, Mark. 2026b. "Beyond Content and Confidence: Constraint Geometry in Belief Arbitration." Manuscript under review.

Watkins, Christopher J. C. H., and Peter Dayan. 1992. "Q-Learning." *Machine Learning* 8 (3): 279–92. https://doi.org/10.1007/BF00992698.